\documentclass{article}

\PassOptionsToPackage{numbers, compress}{natbib}

     \usepackage[preprint]{neurips_2019}
\vbadness=\maxdimen
\usepackage[utf8]{inputenc} 
\usepackage[T1]{fontenc}    
\usepackage{hyperref}       
\usepackage{url}            
\usepackage{booktabs}       
\usepackage{amsfonts}       
\usepackage{nicefrac}       
\usepackage{microtype}      
\usepackage{amsmath}
\usepackage{multirow}
\usepackage{graphicx}
\usepackage{epstopdf}
\let\sss = \scriptscriptstyle

\title{Adversarial Sub-sequence for Text Generation}

\author{
  Xingyuan Chen\textsuperscript{1,3*}, Yanzhe Li\textsuperscript{2}\thanks{Indicates first authors. This work was done while Yanzhe Li was visiting Leshan Normal University.}, Peng Jin\textsuperscript{3\dag}, \\\textbf{Jiuhua Zhang}\textsuperscript{3}, \textbf{Xinyu Dai}\textsuperscript{1\dag}, \textbf{Jiajun Chen}\textsuperscript{1}\thanks{Indicates corresponding authors.}, \textbf{Gang Song}\textsuperscript{4} \\
  $^1$Nanjing University, $^2$Politehnica University of Bucharest, \\$^3$Leshan Normal University, $^4$Henan Leyo Intelligent Technology Company\\
  \texttt{jandp@pku.edu.cn} \\
}

\begin{document}

\maketitle

\begin{abstract}
  Generative adversarial nets (GAN) has been successfully introduced for generating text to alleviate the exposure bias. However, discriminators in these models only evaluate the entire sequence, which causes feedback sparsity and mode collapse. To tackle these problems, we propose a novel mechanism. It first segments the entire sequence into several sub-sequences. Then these  sub-sequences, together with the entire sequence, are evaluated individually by the discriminator. At last these feedback signals are all used to guide the learning of GAN. This mechanism learns the generation of both the entire sequence and the sub-sequences simultaneously. Learning to generate sub-sequences is easy and is helpful in generating an entire sequence. It is easy to improve the existing GAN-based models with this mechanism. We rebuild three previous well-designed models with our mechanism, and the experimental results on benchmark data show these models are improved significantly, the best one outperforms the state-of-the-art model.\footnote[1]{All code and data are available at \url{https://github.com/liyzcj/seggan.git}}
\end{abstract}

\section{Introduction}

Reasonable and meaningful text generation is an important part of many applications such as machine translation \cite{Wu2016Google, bahdanau2014neural}, question answer system \cite{Li2017Adversarial} and image caption \cite{Liu2017Auto, xu2015show}. Neural language model (NLM) \cite{mikolov2010recurrent}, such as Long Short-Term Memory (LSTM) \cite{hochreiter1997long}, have shown excellent performance in text generation. But it will raise exposure bias \cite{Bengio2015Scheduled, ranzato2015sequence}. Generative Adversarial Nets (GAN) \cite{Goodfellow2014Generative} is recently adapted for attacking this issue \cite{Rajeswar2017Adversarial, Lin2017Adversarial, Zhang2017Adversarial}. Unfortunately, the discrete nature of language resulting in that the guild signal from discriminator \(D\) can not be passed back to generator \(G\) through gradient-based method, i.e. non-differentiability issue.

There are mainly two ways to solve the non-differentiability issue. The first way combines reinforcement learning (RL) \cite{williams1992simple} with GAN, the generative model is treated as an agent of RL. The representative models are SeqGAN \cite{Yu2016SeqGAN}, LeakGAN \cite{Guo2017Long} and MaskGAN \cite{Fedus2018MaskGAN} etc. The second way uses a continuous approximate function or continuous latent space to enable the gradient to propagate back\cite{Jang2016Categorical, Maddison2017The}. The representative model is RelGAN \cite{Nie2019ICLR}.

These GAN-based models suffer from mode collapse \cite{Semeniuta2018On}, a crucial reason is lack of informative guiding signals from discriminator \cite{Nie2019ICLR}. Some methods have been proposed to address this issue. \cite{XuDiversity} assigns the novel sentences higher scores than those repeatedly generated ones. \cite{Zhan2018Toward} employs inverse reinforcement learning to optimize policy to maximise the expected total reward. \cite{Nie2019ICLR} feeds multiple embedded representations in the discriminator to provide a more informative signals for the generator training. The mode collapse is alleviated by these ways in a certain extent.

All of above models just evaluate the entire sentence, thus the quality signal of the sub-sentence can only be obtained through the evaluation on the whole sentence. And there are combinatorial explosion and long-distance dependency during the generating sentence, this approach results in the feedback signal about the generation of sub-sentences is sparse. Thus it is difficult to well generate the sub-sentences. The quality of entire sentence is directly related to all its sub-sentences. We could obtain more feedback signals if the whole sentence is evaluated and these sub-sequences are evaluated meanwhile. These signals can improve the quality of generating sub-sentences thus benefit to generate entire sentence. What's more, because sub-sentences are shorter and contains less modes than the whole sentence, thus their distributional are easier to be learned. By evaluating the sub-sentences and entire sentence at the same time, these informative guiding signals will alleviate the mode collapse and improve the quality of the generated sentences.

Therefore, we break the limitation of the discriminator only evaluating on the entire sequence. A novel mechanism is proposed, whereby the entire sequence is segmented into several sub-sequences. All of them, together with the entire sequence, are evaluated by the discriminator individually. Finally, these feedback signals are used to guide the learning of the generator. For an instance, given a sentence \(s=\{w_1w_2w_3w_4\}\), where \(w_i\) is a word, we segment it into three sub-sequences, i.e. \(sub_1=\{w_1\}\),\(sub_2=\{w_1w_2\}\),\(sub_3=\{w_1w_2w_3\}\). All of them together with \(s\) are evaluated by the discriminator individually. We can obtain four guiding signals to update GAN.

Our mechanism has four advantages: (1) more feedback signals. Many sub-sequences are evaluated with the entire sequence individually. (2) more directly feedback signals. These signals directly come from the discriminator, the accuracy will be higher to evaluate the shorter sub-sequence than the entire sequence.  (3) easier to be learned. The shorter sentences are the better to learn their distribution because they contain less modes than the longer sentences. (4) alleviating long-distance dependency. Learning sub-sequences is helpful in learning the entire sequence.

The contributions of this paper are summarized as follows.
\begin{itemize}
\item A novel mechanism is proposed. It makes the adversarial learning not only on the entire sequence but also on the sub-sequences.

\item This novel mechanism can be easily implemented  on the existing GAN-based methods.

\item On benchmark data-sets, we outperforms state-of-the-art model significantly.
\end{itemize}
\section{Related Work}

SeqGAN \cite{Yu2016SeqGAN} first combines reinforcement learning with GAN for text generation. By applying policy gradient \cite{sutton2000policy} method, it optimizes the LSTM generator with rewards received through Monte Carlo (MC) sampling. The reward received by this method has a big gradient variance and the binary rewards is sparse. MaliGan \cite{Tong2017Maximum} trains a model with a maximum likelihood objective to address the issue. RankGan \cite{Lin2017Adversarial} replaces \(D\) with a rank-based model to alleviate sparse guide signal. Leakgan \cite{Guo2017Long} uses a hierarchical reinforcement model with policy gradient. To counter the sparsity issue, they leak internal features from discriminator to obtain more guide signals from \(D\). MaskGAN \cite{Fedus2018MaskGAN} only trains a generator on one sub-sequence to achieve precise rewards. Different from current evaluation mechanism used in reinforcement learning, our mechanism do not only evaluate the whole sequence, but also evaluate the depended sub-sequences and return more useful and dense guide signals to train the generator.

The RL-free model contains applying continuous approximating softmax function and working on latent continuous space directly. GSGAN \cite{Jang2016Categorical} applies  Gumbel-Softmax trick to approximate softmax function. TextGAN adds Maximum Mean Discrepancy to the original objective of GAN based on feature matching \cite{salimans2016improved}. Specifically, FM-GAN \cite{chen2018adversarial}  matches the latent feature distributions of real and synthetic sentences via using a novel metric. RelGAN \cite{Nie2019ICLR} uses a  relational memory-based generator \cite{santoro2018relational}, and employs a multi-head mechanism \cite{Vaswani2017Attention} in the discriminator to prevent the feedback rewards sparsity issue. However all heads in RelGAN’s discriminator received whole sequence. Differently, our mechanism feeds different discriminator with different sub-sequence.

\section{Our model}

In a GAN-based text generation model, the generation is denoted as \(G_\theta\) and a \(\phi\)-parameterized discriminative model denoted as \(D_\phi\).  \(Y_{1:\sss T}=(y_1,y_2,...,y_{\sss T})\) is the sequence generated by  \(G_\theta\), and \(R_{1:\sss T}=(r_1,r_2,...,r_{\sss T})\) is a real sequence. \(R\) is the training set. We denote \(X_{1:\sss T}=(x_1,x_2,...,x_{\sss T})\) as a sequence variable and \(X_{1:\sss T}=(x_1,x_2,...,x_t)\) is one of its sub-sequence, where \(t<=T\). Let \(p_{\sss R}\) be the distributional function on real sequences while \(p_{\theta}\) is the distributional function on the sequences generated by \(G_\theta\). \(p_{\theta}(X_{1:t})\) and \(p_{\sss R}(X_{1:t})\) are the marginal distributions of \(p_{\theta}(X_{1:\sss T})\) and \(p_R(X_{1:\sss T})\) respectively.

\begin{figure}
  \centering
  \includegraphics[scale=0.5]{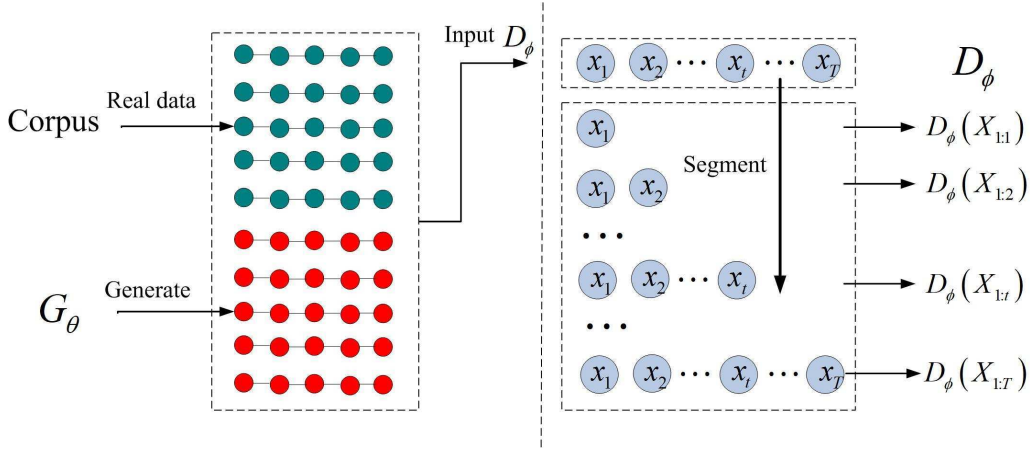}
  \caption{Illustration of segmenting the entire sequence into many sub-sequences.}
  \label{figure1}
\end{figure}

\subsection{Learning to generate sub-sequences is helpful in learning to generate entire sequence}

The GAN is trained according to the divergence between \(p_{\theta}(X_{1:\sss T})\) and \(p_{\sss R}(X_{1:\sss T})\). \(z_{\sss T}\) estimates this divergence, and is shown in the following function.

\begin{equation}\label{1}
z_{\sss T}=D_{\phi}(X_{1:\sss T})
\end{equation}

During the adversarial learning, \(X_{1:\sss T}\) is sampled according to \(G_\theta\) and \(R\). \(z_{\sss T}\) is computed according to the sampling divergence. When \(p_{\theta}(X_{1:\sss T})\neq{p_{\sss R}(X_{1:\sss T})}\), GAN updates \(G_\theta\) and \(D_\phi\) with guide signals \(z_{\sss T}\), make \(p_{\theta}(X_{1:\sss T})\) approximates to \(p_{\sss R}(X_{1:\sss T})\). This is the learning process of \(p_{\theta}(X_{1:\sss T})\).

Given \(T\), the number of all the possible sequence \(X_{1:\sss T}\) is limited. It is easy to proved the marginal distributional \(p_{\theta}(X_{1:t})\) will approximate to \(p_R(X_{1:t})\), when \(p_{\theta}(X_{1:\sss T})\) approximates to \(p_{\sss R}(X_{1:\sss T})\) gradually.

Learning the distribution of sub-sequence is helpful in learning the entire sequence. Because \(p_{\theta}(X_{1:\sss T})=p_{\theta}(X_{1:t})p_{\theta}(X_{t+1:\sss T}|X_{1:t})\), \(p_{\theta}(X_{1:\sss T})\) approximates to \(p_{\sss R}(X_{1:\sss T})\), if and only if \(p_{\theta}(X_{1:t})\) and \(p_{\theta}(X_{t+1:\sss T}|X_{1:t})\) approximate to \(p_{\sss R}(X_{1:t})\) and \(p_{\sss R}(X_{t+1:\sss T}|X_{1:t})\) at the same time. If we can make \(p_{\theta}(X_{1:t})\) approximate to \(p_{\sss R}(X_{1:t})\), then it will be easier to learn \(p_{\theta}(X_{1:\sss T})\). We show this conclusion by the generative process of \(X_{1:\sss T}\). When \(p_{\theta}(X_{1:t})\) approximates to \(p_{\sss R}(X_{1:t})\), \(X_{1:t}\) is generated by \(G_\theta\) according to \(p_{\theta}(X_{1:t})\). So \(X_{1:t}\) will be high quality. It means that the range of quality uncertainty about \(X_{1:\sss T}\) is narrowed from \(1\sim{T}\) to \(t+1\sim{T}\). So, when \(p_{\theta}(X_{1:t})\) approximates to \(p_{\sss R}(X_{1:t})\), the learning process of  \(p_{\theta}(X_{1:\sss T})\) will be easier.

 \(X_{1:t}\) is shorter than \(X_{1:\sss T}\), it contains less modes, thus learning \(p_{\theta}(X_{1:t})\) is easier than learning \(p_{\theta}(X_{1:\sss T})\).

\subsection{Text generation based on sub-sequences}

The way to learn \(p_{\theta}(X_{1:t})\) in GAN is very crucial. In this paper, we propose a novel method which uses \(D_\phi\) to evaluate not only on the entire sequence, but also on sub-sequences.

\begin{equation}\label{2}
z_{t}=D_{\phi}(X_{1:t}), \qquad t=1,2,...,T
\end{equation}

Therefore, we have two kinds of signals, \(z_{\sss T}\) and \(z_t\), to guide the model in learning. During the learning process, it exploits \(z_t\) to make \(p_{\theta}(X_{1:t})\) approximate to \(p_{\sss R}(X_{1:t})\). Obviously, we could obtain \(T-1\) additional updating parameters signals \(z_t\). The distributional functions of short sub-sequences are easier to be learned than the long ones and they are helpful in learning the distribution of the entire sequence. In particular, it will alleviate the long range dependency for generating long sequence.

We use this method to improve GAN. The Figure \ref{figure1} illustrates the new architecture.

Through this method, in addition to the entire sequence, \(D_\phi\) has to predict whether sub-sequences are real or fake. Our experiment shows \(D_\phi\) is qualified to do these multi-task evaluations and improves the quality of the generated texts significantly.

\subsection{Implementation}

Our method can be applied to GAN-based text generation models. In this section, we exemplify this through two different models: SeqGAN, which applies reinforcement learning; and RelGAN, which utilizes a continuous approximation function. The latter achieves the state-of-the-art performance.

\subsubsection{SeqGAN Improvement}
For the SeqGAN, the objective function of discriminator \(D_\phi\) is:
\begin{equation}\label{3}
\mathop{min}\limits_{\phi}-\mathbb{E}_{R_{1:\sss T}\sim{p_R}}\bigg[{log(D_\phi(R_{1:\sss T}))}\bigg]-\mathbb{E}_{Y_{1:\sss T}\sim{p_{\theta}}}\bigg[{1-log(D_\phi(Y_{1:\sss T}))}\bigg]
\end{equation}

In Equation \ref{3}, \(D_\phi\) only evaluates the entire sequence.
The objective function of generator \(G_\theta\) is:
\begin{equation}\label{4}
J(\theta)=\sum_{t=1}^TE_{Y_{1:t-1}\sim{p_{\theta}}}\bigg[\sum_{y_t\in{V}}G_\theta(y_t|Y_{1:t-1})\cdot{Q_{D_\phi}^{G_\theta}(Y_{1:t-1}, y_t)}\bigg]
\end{equation}

The evaluation function for sub-sequence \(Q_{D_\phi}^{G_\theta}(Y_{1:t-1}, y_t)\) is based on Equation \ref{1}.
\begin{equation}\label{5}
Q_{D_\phi}^{G_\theta}(Y_{1:t-1},y_t)=E_{\overline{Y}_{t+1:\sss T}|{Y_{1:t}\sim{p_{\theta}}}}\bigg[D_\phi(Y_{1:t},\overline{Y}_{t+1:\sss T})\bigg]
\end{equation}

Equation \ref{5} is estimated by Monte Carlo search. However, it needs too much computation and may cause a big gradient variance. The expectation in Equation \ref{5} is replaced with \(D_\phi(Y_{1:t})\) based on Equation \ref{2}, we get the Equation \ref{6}.
\begin{equation}\label{6}
Q_{D_\phi}^{G_\theta}(Y_{1:t},y_t)=D_{\phi}(Y_{1:t})
\end{equation}

Compared with the Equation \ref{5}, \(D_\phi\) is only computed once and the evaluation on \(Q_{D_\phi}^{G_\theta}(Y_{1:t-1}, y_t)\) will be exact.
We get the new optimal function for \(D_\phi\) and the new loss function for \(G_\theta\) :

\begin{equation}\label{7}
\mathop{min}\limits_{\phi}-\mathbb{E}_{R_{1:\sss T}\sim{p_{\sss R}}}\bigg[{\sum_{t=1}^{\sss T}log(D_\phi(R_{1:t}))}\bigg]-\mathbb{E}_{Y_{1:\sss T}\sim{p_{\theta}}}\bigg[{\sum_{t=1}^{\sss T}(1-logD_\phi(Y_{1:t}))}\bigg]
\end{equation}

\begin{equation}\label{8}
J(\theta)=\sum_{t=1}^TE_{Y_{1:t-1}\sim{p_{\theta}}}\bigg[\sum_{y_t\in{V}}G_\theta(y_t|Y_{1:t-1})\cdot{D_\phi(Y_{1:t})}\bigg]
\end{equation}

In Equations \ref{7}, \ref{8} By replacing \(Q_{D_\phi}^{G_\theta}(Y_{1:t}, y_t)\) with \(D_\phi(Y_{1:t})\), there is no need for a Monte Carlo search. The reward is directly obtained from \(D_{\phi}\)'s prediction on the sub-sequence \(Y_{1:t}\). We also applied this approach to LeakGAN and our experiment results showed clear improvements.

\subsubsection{RelGAN Improvement}

Similar to the adaption on RL, the original loss function for \(D_{\phi}\) and \(G_{\theta}\) in RelGan will be modified with sub-sequence evaluation function \(D_{\phi}(Y_{1:t})\) individually.

In RelGAN, the discriminator \(D_{\phi}\) is a set of function, \(\{D_{\phi}^{(s)}\}_{s=1}^S\), according to the Equation \ref{1}, where $S$ is the number of discriminator. The loss function is:

\begin{equation}\label{9}
	l_{D}
	=
	\frac{1}{S}
	\sum_{s=1}^S \mathbb{E}_{\mbox{\tiny$\begin{array}{c} R_{1:\sss T}\sim{p_{\sss R}}\\ Y_{1:\sss T}\sim{p_{\theta}}\end{array}$}}f\left( D_{\phi}^{s}( Y_{1:\sss T}),D_{\phi}^{s}( R_{1:\sss T}) \right)
\end{equation}


Using Equation \ref{2} to rewrite the Equation \ref{9}, we get a new loss function:

\begin{equation}\label{10}
	l_{D}
	=
	\frac{1}{S}
	\sum_{s=1}^S \mathbb{E}_{\mbox{\tiny$\begin{array}{c} R_{1:\sss T}\sim{p_{\sss R}}\\ Y_{1:\sss T}\sim{p_{\theta}}\end{array}$}}
		\sum_{t=1}^T f\left( D_{\phi}^{s}( Y_{1:t}),D_{\phi}^{s}( R_{1:t}) \right)
\end{equation}

In Equation \ref{10}, \(D_{\phi}\) will evaluate all sub-sequences.

\subsubsection{Simplified Method}

During the learning process, the marginal distributions of different sub-sequences have their own convergence speeds. It is hard to coordinate these convergence speeds in Equation \ref{7}, \ref{8} and \ref{10}. Meanwhile, there will be so many discriminators that it is hard to be implemented. In this paper, we only want to verify the effectiveness of our mechanism rather than finding the best results based on this mechanism. Therefore, we simplify our method, which is described below:

In Equation \ref{2}, there are \(T\) segments in total. We only select two segments: one is the entire sequence itself \(Y_{1:\sss T}\) and the other is \(Y_{1:T_{ave}}\). \(T_{ave}\) is the average sentences length in the training corpus.

For SeqGAN, the discriminator \(D_{\phi}\) is optimized as follow:
\begin{equation}\label{11}
    \begin{split}
    & \mathop{min}\limits_{\phi}-\mathbb{E}_{R_{1:\sss T}\sim{p_{\sss R}}}\bigg[{logD_\phi(R_{1:T_{ave}})+logD_\phi(R_{1:\sss T})}\bigg]\\
    & -\mathbb{E}_{Y_{1:\sss T}\sim{p_{\theta}}}\bigg[{2-logD_\phi(Y_{1:T_{ave}})-logD_\phi(Y_{1:\sss T})}\bigg]
    \end{split}
\end{equation}
The corresponding objective function of \(G_{\theta}\), \(Q_{D_\phi}^{G_\theta}(Y_{1:t-1}, y_t)\) in the Equation \ref{4} is rewritten as follows:

\begin{equation}\label{12}
    Q_{D_\phi}^{G_\theta}(Y_{1:t-1}, y_t)=
    \begin{cases}
    \frac{1}{N}\sum_{n=1}^ND_{\phi}(Y_{1:T_{ave}}^{n}),\quad  Y_{1:T_{ave}}^{n}\in{MC_1^{G_{\beta}}}(Y_{1:t};N)  \qquad t<T_{ave}\\[2ex]
    \frac{1}{N}\sum_{n=1}^ND_{\phi}(Y_{1:\sss T}^{n}), \quad Y_{1:T}^{n}\in{MC_2^{G_{\beta}}}(Y_{1:t};N) \qquad T_{ave}<t<T\\[2ex]
    \qquad \qquad D_{\phi}(Y_{1:t}) \qquad \qquad  \qquad \qquad \qquad \qquad   t=T_{ave} \quad or\quad T
    \end{cases}
\end{equation}

where \(MC_{1}^{G_{\beta}}(Y_{1:t};N)=\{Y_{1:T_{ave}}^{1},...,Y_{1:T_{ave}}^{N}\}\), \(MC_{2}^{G_{\beta}}(Y_{1:t};N)=\{Y_{1:\sss T}^{1},...,Y_{1:\sss T}^{N}\}\). In the same manner as with SeqGAN, \(Y_{1:t}^n=(y_1,...,y_t)\) and \(Y_{t+1:\sss T}^n\) is sampled based on the roll-out policy \(G_{\beta}\). \(G_{\beta}\) is set the same as the generator \(G_{\theta}\).

Similarly, for improving the RL\(-\)free models, such as RelGAN, we modify the Equation \ref{9} as follows:
\begin{equation}\label{13}
    	l_{D}=\frac{1}{S}
    	\sum_{s=1}^S \mathbb{E}_{\mbox{\tiny$\begin{array}{c}R_{1:\sss T}\sim{p_{\sss R}}\\ Y_{1:\sss T}\sim{p_{\theta}}\end{array}$}}\big[f(D_{\phi}^{s}(Y_{1:T_{ave}}),D_{\phi}^{s}(R_{1:T_{ave}}))+f(D_{\phi}^{s}(Y_{1:\sss T}),D_{\phi}^{s}(R_{1:\sss T}))\big]
\end{equation}

\section{Experiments}
\label{others}

We experiment on three benchmark data sets. One is comprised of synthetic data and the others contain real-world data: COCO image captions dataset and EMNLP2017 WMT news dataset. As with \cite{Nie2019ICLR}, \(NLL_{gen}\) is used for evaluating sample diversity. For evaluating sample quality, \(NLL_{oracle}\) is adapted for synthetic data, while BLEU \cite{Papineni2002BLEU} is used for real scenarios because there is no oracle.

Three very strong baselines are compared. SeqGAN and LeakGAN both use reinforcement learning, and RelGAN makes use of continuous approximation. We adapt the same hyper-parameters setting as the previous models individually. It should be noted that the temperature control is a key hyper-parameter to trade-off the sample quality and diversity in RelGAN.
\begin{equation}\label{14}
NLL_{gen}=-\mathbb{E}_{r_{1:\sss T}\sim{p_{\sss R}}}{logp_{\theta}(r_1,...r_{\sss T})},\quad NLL_{oracle}=-\mathbb{E}_{Y_{1:\sss T}\sim{p_{\theta}}}{logp_{\sss R}(y_1,...y_{\sss T})}
\end{equation}
\subsection{Synthetic Data}

Following \cite{Yu2016SeqGAN, Guo2017Long}, a randomly initialized LSTM with the normal distribution \(\mathcal{N}(0,1)\) as the oracle is used to generate the real data distribution\(G_{oracle}(x_t|x_1,...,x_{t-1})\). 10,000 sequences of length N are generated as the training set \(\mathcal{S}\). In order to verify our model’s performance at different lengths of N, we set the N=20 and N=40 respectively.

\begin{table}[h!]\small
  \setlength{\abovecaptionskip}{0.5cm}
  \setlength{\belowcaptionskip}{0cm}
  \centering
  \begin{tabular}{c|ccccc|c}
    \toprule
    Length & MLE & SeqGAN & LeakGAN & RelGAN & Imp-RelGAN & Real\\
    \midrule
    20 & 9.038 & 8.736 & 7.038 & 6.680 & $\textbf{6.310}\pm{\textbf{0.512}}$ & 5.750 \\

    40 & 10.411 & 10.310 & 7.191 & 6.765 & $\textbf{5.920}\pm{\textbf{0.098}}$ & 4.071\\
    \bottomrule
  \end{tabular}
  \caption{The sample qualify and sample diversity on synthetic data. All the improved models with our novel mechanism are run with five random initialization and other scores are cited directly from their published paper. The "Imp-RelGAN" denotes the improved RelGAN with our novel mechanism. For the \(NLL_{oracle}\) score, the lower the better.}
  \label{synthetic data}
\end{table}

The results are shown in Table \ref{synthetic data}. The improved RelGAN with our novel mechanism achieve the state-of-the art performances on both short and long synthetic sentences. In particular, to the long sentences, it makes even much more progress than the short ones. It shows that this mechanism effectively alleviates the long-distance dependency difficulty in text generation.

\subsection{COCO image captions dataset}

In order to observe the performance on real data, we first select this dataset whose sentences' average length is about 11 words. For comparability, we use the same training and test data as \cite{Nie2019ICLR, Guo2017Long}. There are total 4,682 word types and the longest sentence consists of 37 words. Both the training and test data contain 10,000 sentences.

\begin{table}[ht]\small
  \setlength{\tabcolsep}{1.8mm}
  \setlength{\belowcaptionskip}{0cm}
  \centering
  \begin{tabular}{c|c c c c|c}
    \toprule[2pt]
    Method & BLEU-2 & BLEU-3 & BLEU-4 & BLEU-5 & $NLL_{gen}$\\
    \midrule[1pt]
    MLE & 0.731 & 0.497 & 0.305 & 0.189 & 0.718\\
    \hline
    SeqGAN & 0.745 & 0.498 & 0.294 & 0.180 & 1.082\\

    Imp-SeqGAN & $0.774\pm{0.011}$ & $0.554\pm{0.015}$ & $0.345\pm{0.014}$ & $0.212\pm{0.012}$ & $0.836\pm{0.016}$\\
    \hline
    LeakGAN & 0.746 & 0.528 & 0.355 & 0.230 & 0.679\\

    Imp-LeakGAN & $0.825\pm{0.036}$ & $0.668\pm{0.034}$ & $0.495\pm{0.029}$ & $0.339\pm{0.028}$ & $\textbf{0.584}\pm{\textbf{0.018}}$\\
    \hline
    RelGAN(100) & 0.849 & 0.687 & 0.502 & 0.331 & 0.756\\

    Imp-RelGAN(10) & $\textbf{0.879}\pm{\textbf{0.009}}$ & $\textbf{0.734}\pm{\textbf{0.015}}$ & $\textbf{0.556}\pm{\textbf{0.023}}$ & $\textbf{0.390}\pm{\textbf{0.025}}$ & $0.697\pm{0.015}$\\
    \hline
    RelGAN(1000) & 0.814 & 0.634 & 0.455 & 0.303 & 0.655\\

    Imp-RelGAN(50) & $0.845\pm{0.011}$ & $0.676\pm{0.018}$ & $0.484\pm{0.023}$ & $0.320\pm{0.024}$ & $0.615\pm{0.012}$\\
    \bottomrule[2pt]
  \end{tabular}
  \caption{The sample quality and sample diversity on COCO Image Caption. All the improved models with our novel mechanism are run with five random initialization and other scores are cited directly from their published paper. The numbers in parentheses are the temperature for all kinds of RelGAN. Imp-X denotes the improved model X with our novel mechanism.}
   \label{coco_image}
\end{table}

The results can be seen in Tabel \ref{coco_image}. When the SeqGAN is modified with our mechanism, all of its BLEU scores increase and \(NLL_{gen}\) decreases. It means that its sample quality and diversity are improved. The LeakGAN is in the same situation. To RelGAN, its performance is closely related to the temperature.  When the temperature is decreased, the sample quality improves but the sample diversity declines. Given any temperature $t$, the sample quality  and the sample diversity of the model cannot be exceeded simultaneously at any other temperature. For the RelGAN that is modified with our mechanism, with temperature 10, the sample quality and diversity are improved significantly at the same time. It is in a similar situation to other temperatures. The improved RelGAN outperforms the previous RelGAN and achieves the state-of-the-art performance.

\subsection{EMNLP2017 WMT news dataset}

 In this dataset, the average length of sentences is about 20 words. There are total 5,255 word types and the longest sentence is consisted of 51 words. Similar to COCO, we directly use directly the training and test data from Texygen \cite{Zhu2018Texygen} . All training data is used\footnote[2]{We contacted with the first author of RelGAN, he said there was a tpyo error in his paper and he used all training data.}. There are 10,000 sentences in test data.

\begin{table}\small
  \setlength{\tabcolsep}{1.8mm}
  \setlength{\belowcaptionskip}{0cm}
  \centering
  \begin{tabular}{c|c c c c|c}
    \toprule[2pt]
    Method & BLEU-2 & BLEU-3 & BLEU-4 & BLEU-5 & $NLL_{gen}$\\
    \midrule[1pt]
    MLE & 0.768 & 0.473 & 0.240 & 0.126 & 2.382\\
    \hline
    SeqGAN & 0.777 & 0.491 & 0.261 & 0.138 & 2.773\\

    Imp-SeqGAN & $0.778\pm{0.008}$ & $0.493\pm{0.006}$ & $0.263\pm{0.009}$ & $0.140\pm{0.009}$ & $2.547\pm{0.164}$\\
    \hline
    LeakGAN & 0.826 & 0.645 & 0.437 & 0.272 & 2.356\\

    Imp-LeakGAN & $0.882\pm{0.002}$ & $0.710\pm{0.003}$ & $0.486\pm{0.003}$ & $0.292\pm{0.001}$ & $2.344\pm{0.013}$\\
    \hline
    RelGAN(100) & 0.881 & 0.705 & 0.501 & 0.319 & 2.482\\

    Imp-RelGAN(10) & $\textbf{0.893}\pm{\textbf{0.004}}$ & $\textbf{0.728}\pm{\textbf{0.008}}$ & $\textbf{0.516}\pm{\textbf{0.011}}$ & $\textbf{0.322}\pm{\textbf{0.011}}$ & $2.272\pm{0.025}$\\
    \hline
    RelGAN(1000) & 0.837 & 0.654 & 0.435 & 0.265 & 2.285\\

    Imp-RelGAN(50) & $0.880\pm{0.007}$ & $0.693\pm{0.012}$ & $0.469\pm{0.016}$ & $0.282\pm{0.013}$ & $\textbf{2.165}\pm{\textbf{0.014}}$\\
    \bottomrule[2pt]
  \end{tabular}
  \caption{The sample quality and sample diversity on EMNLP2017 WMT News. All the improved models with our novel mechanism are run with five random initialization and other scores are cited directly from their published paper. The numbers in parentheses are the temperature for all kinds of RelGAN. Imp-X denotes the improved model X with our novel mechanism.}
  \label{emnlp}
\end{table}

Table \ref{emnlp} gives the results. Similar to COCO image captions dataset, all models are modified with our mechanism outperforms the previous counterparts on this dataset. This demonstrates that our method works still well on long sentences.


A Turing test is performed for the generated sentences. A person assigns one sentence zero score if he thinks it is generated by machine otherwise assigns it one credit. In order to evaluate one model, he will be provided 100 sentences simultaneously, half of them are real and the rest are randomly selected from the generated sentences by this model. We evaluate all models one by one. 10 university students majoring in English, score every sentences. The experiment results are listed in Table \ref{Turing test}. It indicates the generated sentences of the modified RelGAN are better than other models.

\begin{table}[h]\small
  \setlength{\tabcolsep}{1.8mm}
  \setlength{\belowcaptionskip}{0cm}
  \centering
  \begin{tabular}{c|c c c c c|c}
    \toprule
    Method & MLE & SeqGAN & LeakGAN & RelGAN & Imp-RelGAN & Real\\
    \midrule
    Human Score & $0.21\pm{0.10}$ & $0.28\pm{0.21}$ & $0.36\pm{0.10}$ & $0.32\pm{0.04}$ & ${\textbf{0.54}\pm{\textbf{0.14}}}$ & $0.72\pm{0.08}$\\
    \bottomrule
\end{tabular}
\caption{The Turing test results. "Real" denotes the human score on the real data.}
\label{Turing test}
\end{table}

\section{Analysis and Case Study}

\begin{figure}
  \setlength{\abovecaptionskip}{0cm}
  \setlength{\belowcaptionskip}{0cm}
  \centering
  \includegraphics[scale=0.3]{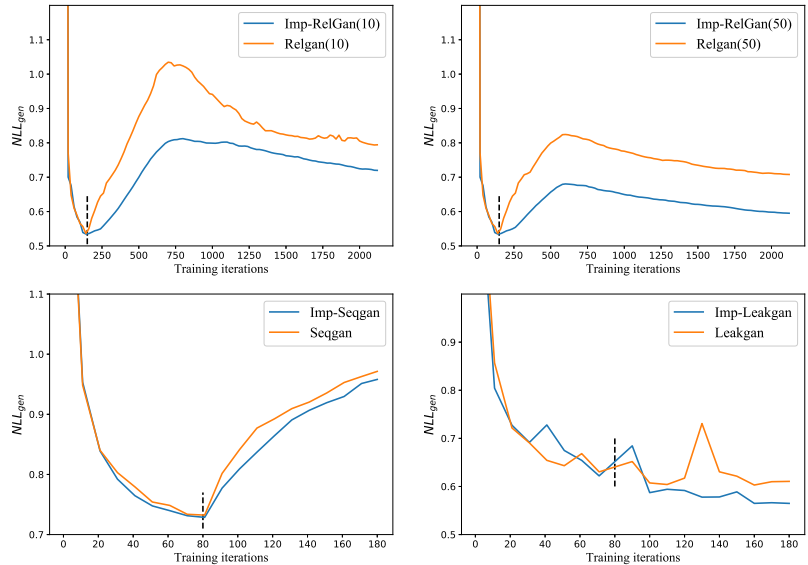}
  \caption{Training curves of $NLL_{gen}$ on COCO Image Captions with different models: RelGAN(10), RelGAN(50), SeqGAN, LeakGAN. Imp-X denotes the improved model X with our novel mechanism. We can see that the $NLL_{gen}$ of improved models are consistently better than origin models, which demonstrates the advantages of our mechanism.}
  \label{figure2}
\end{figure}

The learning curves of $NLL_{gen}$ compared with different models are provided in Figure \ref{figure2}. Note that at the same temperature, the BLEU scores of RelGAN and Imp-RelGAN are very closed to each other, but the $NLL_{gen}$ of Imp-RelGAN is much lower than RelGAN during the whole training. What'more, the BLEU scores of Imp-LeakGAN and Imp-SeqGAN are higher than LeakGAN and SeqGAN. It reveals that our mechanism can alleviate mode collapse meanwhile remain the sample quality.

\begin{table}[ht!]\small
    \setlength{\abovecaptionskip}{0.5cm}
    \setlength{\belowcaptionskip}{0cm}
    \centering
    \begin{tabular}{p{2cm}p{5cm}p{5cm}}
        \toprule
        Datasets & RelGan & Imp-RelGan  \\
        \midrule
        \multirow{2}*{\shortstack{COCO Image \\Captions}} & (1) a white toilet sits on the side of a toilet in a bathroom .  & (1) a young boy stands next to a row of parked motorcycles in a parking lot . \\ & (2) a man standing next to sheep in a street next to a parking garage . & (2) a black and white dog sitting in the basket of a bicycle . \\
        \midrule
        \multirow{2}*{\shortstack{EMNLP2017 \\ WMT}} & (1) the human body of state 's for the first time in the year , most of the government ' s long - time to the individual . & (1) " i have to think about the freedom of expression and the way i ' m performing , " he told the french people . \\ & (2) this is time for scotland to come in and not to the majority of eu voters who are not in control of the law . & (2) our older players are starting to understand that we don ' t always get the chance to go on and answers to them . \\
        \bottomrule
    \end{tabular}
    \caption{Samples from different methods on COCO Image Captions and EMNLP2017 WMT News.}
    \label{samples}
\end{table}

A few samples generated by RelGAN and its modified version with our novel mechanism are shown in Table \ref{samples}. More Samples are provided in the supplementary material.

\section{Conclusion}

In this paper, we propose a novel mechanism for GAN to evaluate the entire sequence and the sub-sequences segmented from it, rather than just evaluating the entire sequence.  Experiments on both synthetic data and two real benchmark data-sets show our mechanism works very well on three GAN-based models. All of them are improved significantly and the best one achieve the state-of-the-art sample quality and sample diversity.

A natural extension to our mechanism is applying our method for image generation. Secondly, the importance of mode collapse \cite{Tong2017Maximum, Tong2016Mode, Srivastava2017VEEGAN}, we will observe its variations with the number of the sub-sequences. At last, we will try to design a new method to adjust the temperature parameter in order to balance the sample quality and diversity.

\bibliographystyle{abbrv}
\bibliography{cites}
\end{document}